\documentclass[conference]{IEEEtran}

\usepackage{multirow}
\usepackage{graphicx}
\usepackage{mathtools}
\usepackage{amsmath}
\usepackage{url}
\usepackage{booktabs}
\usepackage{csquotes}
\usepackage{bm}
\usepackage{algorithm}
\usepackage{algpseudocode}
\usepackage{multicol}

\usepackage{amsmath,amssymb,amsfonts}

\DeclarePairedDelimiter\autobracket{(}{)}
\newcommand{\br}[1]{\autobracket*{#1}}
\newcommand{\R}{\mathbb{R}}
\newcommand{\F}{\mathcal{F}}
\newcommand{\eps}{\varepsilon}
\newcommand{\E}{\mathbb{E}}
\newcommand{\N}{\mathcal{N}}
\newcommand{\LB}{\mathcal{L}}
\newcommand{\q}{q\brt}
\DeclareMathOperator*{\argmin}{arg\,min}
\DeclarePairedDelimiter\sparen{[}{]}
\newcommand{\sbr}[1]{\sparen*{#1}}

\newcommand{\iI}{\mathcal{I}^{-1}}
\newcommand{\natgrad}{\tilde{\nabla}}
\renewcommand{\S}{\Sigma}

\newtheorem{prop}{Proposition}
\renewcommand{\it}{{t^{-1}}}

\newcommand{\bmu}{{\bm{\mu}}}
\newcommand{\bth}{{\bm{\theta}}}
\newcommand{\bz}{{\bm{\zeta}}}
\newcommand{\bnu}{{\bm{\nu}}}
\newcommand{\brt}{\br{\bm{\theta}}}

\usepackage[numbers]{natbib}

\usepackage{hyperref}

\begin{document}

\title{Variational Inference for GARCH-family Models}

\author{\IEEEauthorblockN{Martin Magris, Alexandros Iosifidis}
\IEEEauthorblockA{\textit{Department of Electical \& Computer Engineering} \\
\textit{Aarhus University}\\
Aarhus, Denmark \\
\{magris, ai\}@ece.au.dk}
}

\maketitle

\begin{abstract}
The Bayesian estimation of GARCH-family models has been typically addressed through Monte Carlo sampling. Variational Inference is gaining popularity and attention as a robust approach for Bayesian inference in complex machine learning models; however, its adoption in econometrics and finance is limited. This paper discusses the extent to which Variational Inference constitutes a reliable and feasible alternative to Monte Carlo sampling for Bayesian inference in GARCH-like models. Through a large-scale experiment involving the constituents of the S\&P 500 index, several Variational Inference optimizers, a variety of volatility models, and a case study, we show that Variational Inference is an attractive, remarkably well-calibrated, and competitive method for Bayesian learning.

\end{abstract}


\section{Introduction}\label{sec:introd}
The classical estimation procedures for GARCH-family modes are the frequentist maximum likelihood, quasi maximum likelihood, and the generalized method of moments approaches \citep{bollerslev1992arch}. Recently, there has been a growing interest in using Bayesian estimation techniques, as they offer several advantages over the traditional approaches \citep{ardia2010efficient}. For instance, in the frequentist approach, models are compared with no other means that their likelihood, whereas Bayes factors and marginal likelihood allow comparisons of non-nested models. Bayesian estimation provides reliable results in finite samples, and, e.g., can uncover the full value-at-risk density. 
Maximum likelihood estimators present some limitations when the errors are heavy-tailed and may not be asymptotically Gaussian \citep{hall2003inference}, and positivity of the conditional variance and stationarity requirements can lead to complex non-linear inequalities, making constrained optimization cumbersome. 
For the Bayesian estimation of GARCH-family models, Monte Carlo (MC) sampling is the predominant approach \cite[e.g.,][]{hall2003inference,ardia2008financial,virbickaite2015bayesian}. Indeed the recursive nature of the conditional variance makes the joint posterior of unknown parametric form, and the choice of the sampling algorithm is crucial. The Griddy-Gibbs sampler has extensively been used in this context  \citep[e.g.,][]{bauwens1998bayesian,ausin2007bayesian}, along with importance sampling \citep[e.g.,][]{geweke1989bayesian,kleibergen1993non}, and the Metropolis-Hastings (MH)
\citep{muller1998monte,gerlach2008bayesian}. Different extensions of the MH algorithm have been proposed \citep[e.g.,][]{nakatsuma1998markov}, along with the use of alternative methods \citep[e.g.,][]{ardia2009bayesian}. For a broader overview,  see, e.g., the surveys \citep{ardia2010efficient,virbickaite2015bayesian}, or the textbook \citep{ardia2008financial}. 

The ability of Bayesian methods to address uncertainty via posterior distribution gained much attention in Machine Learning (ML). In the last decade, sophisticated Bayesian methods have been advanced for training high-dimensional ML models, and the theory of Bayesian neural networks has been extensively developed \citep[see, e.g.,][for a review]{magris2023survey}. Under the complexity of typical ML models, sampling methods do not scale up in high dimensions and difficult to apply. Variational Inference (VI) stands as a successful and feasible alternative, widely exploited in ML applications \citep{kingma2013auto,ranganath2014black,kucukelbir2017automatic}. VI reduces the typical Bayesian integration problem to a simpler optimization problem aimed at finding the \enquote{best} approximation of the true posterior distribution in the sense described in Sec. \ref{sec:vi}. Despite its consolidated use in ML, VI has not received much attention in econometrics and finance as a feasible Bayesian alternative to MC sampling.

In particular, the use of VI as a tool for the Bayesian estimation of GARCH-family models remains unaddressed. Though VI has been used in volatility forecasting with ML models \citep{nguyen2022recurrent,Nguyen2019JBES},
there have been few self-contained VI applications concerning GARCH models \citep{magris2022quasi,magris2022exact,tran2021variational}.
This paper fills the this gap and addresses the feasibility and appropriateness of VI as an alternative to MC sampling and maximum likelihood estimation by conducting extensive experiments on the constituents of the S\&P500 index. We show how Gaussian VI can be implemented and applied at a large scale as an unconstrained optimization problem through appropriate parameter transforms. We validate the results over several in-sample and out-of-sample performance metrics. Along with a focused study on the Microsoft Inc. stock data, with different optimization algorithms, we show VI is an effective, robust, and competitive approach for the Bayesian estimation of various GARCH-family models.

\section{Methods}
\subsection{GARCH models} \label{sec:models:garch}
A major task of financial econometrics is modeling volatility in asset returns.
Volatility is considered a measure of risk for which investors
demand a premium for investing in risky assets. Empirical observations of financial returns reveal some stylized facts. Whereas returns are nearly uncorrelated, they contain higher-order dependences. The correlation of absolute and squared returns is positive and persistent. Autocorrelated daily volatility thus appears to be predictable, and  Autoregressive Conditional Heteroskedasticity (ARCH) models provide the basis for the most popular parameterizations of this dependence.

Let $\eps_t$ be a random variable that has mean and variance conditionally on the information set $\F_{t-1}$ ($\sigma$-algebra generated by $\eps_{t-j},j \geq 1$). For the ARCH model, $\E\br{\eps_t\vert \F_{t-1}}=0$ and the conditional variance $h_t=\E\br{\eps^2_t\vert \F_t}$ is a parametric function on $\F_{t-1}$ \citep{terasvirta2009introduction}. The sequence $\{\eps_t\}$ may be observed or be an error innovation sequence of an econometric model: $\eps_t = r_t-\mu_t\br{r_t}$ with $r_t$ and observable random variable (e.g., a daily return) and $\mu_t\br{r_t}$ the conditional expectation of $r_t$ given $\F_{t-1}$. The parametric form of the ARCH model reads:
\begin{align*}
    r_t &= \mu + \eps_t,\\
    \eps_t &= h^{1/2}_t z_t\text{,} \quad z_t \sim \text{\textit{iid }} \N\br{0,1},\\
    h_t &= \omega + \alpha_1 \eps^2_{t-1},
\end{align*}
with $t=1,\dots,T$ and, to ensure $h_t>0$ and identification, $\omega >0$, $0< \alpha_1 <1$. We shall assume $\mu\equiv0$. $h_t$ is the conditional and time-dependent volatility of $\eps_t$. The way  $\eps_t$ is defined guarantees white noise properties, since $z_t$ is a sequence of iid variables.
Normality is a typical assumption for the iid sequence $\{z_t\}$, but leptokurtic alternatives are also used. The iid assumption guarantees the white noise property of $\{\eps_t\}$.
In the ARCH equation defining the parametric form for the conditional variance, the linear function of the squared innovation at $t-1$ can be generalized to a higher-order ARCH$(q)$:
$$
h_t = \omega + \sum_{j=1}^q \alpha_j \eps^2_{t-j},
$$
where $\omega>0,\, \alpha_j \geq 0$, with at least an $\alpha_j>0$. Note that the volatility, the object of modeling, is not observed: using $\eps^2_t$ is an immediate solution, but alternatives exist if, e.g., the data is available at intraday frequencies.

For the ARCH family, the decay rate in the autocorrelation of $\eps_t^2$ is too rapid compared to the observed time series: the so-called Generalized ARCH (GARCH) is a predominant alternative. In a GARCH$(p,q)$ model, the conditional variance is not only a function of the lagged innovations but also of its lags:
$$
h_t = \omega + \sum_{j=1}^q \alpha_j \eps^2_{t-j} + \sum_{j=1}^p \beta_j h_{t-j}.
$$
The overwhelming model has been the GARCH$(1,1)$. Sufficient conditions for the positivity of conditional variances are  $\omega >0$, $\alpha_j \geq 0$, $j=1,\dots,q$ and $\beta_j \geq 0$, $j=1,\dots,p$. Identifiability requires at least one $\beta_j>0$ and one $\alpha_j>0$, and for stationarity $\sum \alpha_j + \sum \beta_j <1$.

GARCH models have been extended and generalized in many different directions. Among these, the empirical evidence of asymmetry in volatility clustering motivates the GJR-GARCH \citep{glosten1993relation} model, which assumes the response of the variance to a shock not to be independent of its sign:
$$
h_t = \omega + \sum_{j=1}^q  \alpha_j  + \sum_{j=1}^o \gamma_j I\br{\eps_{t-j}>0} \eps^2_{t-j} + \sum_{j=1}^p \beta_j h_{t-j},  
$$
with $ I\br{\cdot}$ an indicator function, defines the GJR-GARCH$(p,o,q)$  (with $o=0$ we simply write GJR-GARCH$(p,q)$). It must hold $\omega>0$, $\alpha_j\geq 0$, $\beta_j  \geq 0$, $\gamma_j \geq0$, $\sum \alpha_j+ \gamma_j\geq 0$, and $\sum \alpha_j + 1/2\sum \gamma_j + \sum \beta_j<1$.

The Exponential GARCH (EGARCH) model is another popular extension. The family of EGARCH$(p,q)$ models can be defined with
$$
\log h_t = \omega + \sum_{j=1}^q g_j\br{z_{t-j}} + \sum_{j=1}^p \beta_j \log h_{t-j}.
$$
In our analyses we adopt the version of \citep{nelson1991conditional} where $g_j\br{z_{t-j}} = \alpha_j z_{t-j} + \psi_j\br{\vert z_{t-j}\vert -\E\br{\vert z_{t-j}\vert }}$. The model does not impose any restriction on the parameters because, since the equation is on log variance instead of variance itself, the positivity of the variance is automatically satisfied. This is a big advantage in model estimation. For a concise presentation of the advantages and limitations of the EGARCH model, refer, e.g., to \cite{terasvirta2009introduction}. By including $\gamma_j I\br{\eps_{t-j}>0} \eps^2_{t-j}$, $j=1,\dots,o$ terms in the above conditional variance equation, one defines the GJR-EGARCH$(p,o,q)$.

In our analyses, we furthermore adopt the Fractionally Integrated GARCH (FIGARCH). The FIGARCH model \citep{baillie1996fractionally} conveniently
explains the slow decay in autocorrelation functions of squared observations of typical daily return series. With the FIGARCH, the effect of the lagged $\eps^2_t$
on $h_t$ decays hyperbolically as a function of the lag length. The FIGHARCH$(p,d,m)$ process is defined as:
$$
\br{1-L}^d\phi\br{L}\eps^2_t = \bar{\omega} + \br{1-\beta\br{L}}v_t \text{,}
$$
where $L$ is the lag operator, $\phi\br{L} =\sum_{j=1}^{m-1} \phi_jL^j$, $\beta\br{L} = \sum_{j=1}^p \beta_j L^j$, $v_t = \eps^2_t -h^2_t$, and $d$ is the order of fractional differencing that guides the long-memory properties of the process \citep{terasvirta2009introduction}. 
Of relevance for estimation is its equivalent ARCH($\infty$) representation of the model:
\begin{equation}\label{eq:figarch}
 h_t = \omega + \sum_{j=1}^\infty \lambda_k\eps^2_{t-k}\text{,}   
\end{equation}
where $\omega>0$, and $\lambda_k \geq 0$ are recursively defined. For the FIGARCH $(1,d,1)$, $\delta_1 = d$, $\lambda_1 = \phi -\beta + d$, $\delta_k = (k-1-d)/k\delta_{k-1}$, $\lambda_k = \beta \lambda_{k-1} + \delta_k -\phi\delta_{k-1}$, with the constrains $\omega>0$, $0\leq d \leq 1$, $0 \leq \phi \leq (1-d)/2$, $0 \leq \beta \leq d+ \phi$, sufficient to ensure the positivity of the conditional variance \citep{baillie1996fractionally}.

\subsubsection{Estimation}
With a possibly misspecified but convenient standard likelihood function and the assumption that the dynamic of the volatility process is correctly specified, the models described earlier are generally estimated via Quasi Maximum Likelihood (QML).
Under a Gaussian likelihood, the QML objective generally reads:
\begin{equation} \label{eq:garch_qml}
   \ell \br{\bnu} = \sum_{t=1}^T \br{ \log h_t (\bnu) + \frac{\eps^2_t}{h_t(\bnu)} }\text{,}
\end{equation}
where $\bnu$ collects all the relevant parameters, e.g., for the GARCH(1,1), $\bnu = \br{\omega,\alpha_1,\beta_1}$, and the dependence of the conditional variance on it, is made explicit. Constrained gradient descent procedures are effective for minimizing \eqref{eq:garch_qml}. Sec. \ref{sec:trans} discusses using parameter transforms to perform unconstrained optimization. Eq. \eqref{eq:garch_qml} implies a recursive relation whose implementation is expensive. For initialization, it is common to back-cast $\max\{p,o,q\}$ values with the average value of $\{r^2_t\}$.

\subsection{Variational Inference} \label{sec:vi}
\subsubsection{General principle}
Let $y$ denote the data and $p\br{y\vert \bth}$ the likelihood of the data based on a postulated model with $\bth \in \Theta$ a $d$-dimensional vector of model parameters. Let $p\brt$ be the prior distribution on $\bth$. The goal of Bayesian inference is the posterior distribution 
 $p\br{\bth\vert y} = p\br{y,\bth}/p\br{y}$,
where $p\br{y}=\int_\Theta  p\br{y\vert \bth} p\brt d\bth$. 
Bayesian inference is generally difficult since the marginal likelihood  $p\br{y}$ is often intractable and of unknown form, and 
Variational Inference (VI) is an attractive alternative.

VI consists of an approximate method approximating the posterior distribution by a probability density $\q$ (called variational distribution) belonging to some tractable class of distributions $\mathcal{Q}$. VI thus turns the Bayesian inference problem into that of finding the best approximation $q^\star \brt \in \mathcal{Q}$ to $p\br{\bth\vert y}$ by minimizing the Kullback-Leibler (KL) divergence from $\q$ to $p\br{\bth\vert y}$:
\begin{equation*}
    q^\star = \argmin_{q \in \mathcal{Q}}  \text{KL}\br{q \vert \vert p\br{\bth\vert y}}   = \argmin_{q \in \mathcal{Q}} \int \q \log \frac{\q}{p\br{\bth\vert y}} d\bth \text{.}
\end{equation*}
It can be shown the KL minimization is equivalent to the maximization of the so-called Lower Bound (LB) on $\log p\br{y}$, \citep[e.g.][]{tran2021variational}:
\begin{equation}\label{eq:lb}
    \LB\br{q} \coloneqq \int \q \log \frac{ p\br{y\vert \bth} p\brt}{\q} d \bth 
    = \E_q \sbr{h\brt} \text{,}
\end{equation}
with $h_\bz\brt = \log p(y\vert \bth) + \log p(\bth) -\log q_\bz \brt$.
In fixed-form VI, the parametric form of the variational posterior is set. Typically, the target is a Gaussian distribution of mean $\bmu$ and covariance $\S$, and $q_\bz$ in the set $\mathcal{Q}$ of Gaussian distributions, with $\bz = \{\bmu,\text{vec}(\S)\}$ a vector of parameters. VI seeks the parameter $\bz^\star$ optimizing \eqref{eq:lb}.

The standard approach for maximizing the LB is based on a stochastic gradient descent update, whose basic form is
\begin{equation} \label{eq:sgd_on_zeta}
    \bz_{t+1} = \bz_t+\delta \left. \sbr{\iI_\bz \hat{\nabla}_\bz\mathcal{L}(q_\bz) } \right|_{\bz = \bz_t} \text{,}
\end{equation}
where $t$ denotes the iteration, $\delta$ a the step size, and $\hat{\nabla}_\bz \LB(q_\bz)$ a stochastic estimate of the Euclidean gradient. 
In place of Euclidean gradients, the recent literature adopts natural gradients leading to improved step directions by accounting for the information geometry of the variational distribution \citep[see, e.g.,][]{magris2023survey}. With natural gradients, $\iI_\bz$ is the corresponding Fisher Information Matrix, otherwise is the identity matrix $I$, of size equal to the number of trainable parameters $d$.

\subsubsection{Algorithms}
A major aspect of implementing \eqref{eq:sgd_on_zeta} is the gradient computation. Methods requiring the actual computation of the gradients of the loss, such as the reparametrization trick \citep{kingma2013auto}, are expensive to implement at a large scale within the recurrent form of the likelihood \eqref{eq:garch_qml}.
Furthermore, the use of automatic differentiation is not a widespread practice in econometrics and finance, largely adopting numerical differentiation. The approaches discussed here rely on the use of the log-derivative trick for evaluating the gradient of the expectation $\E_{q_\bz}\sbr{h_\bz\brt}$ as an expectation of a gradient:
\begin{align}\label{eq:bb_gradient}
    \natgrad_\bz\LB(q_\bz) =  
    \iI_\bz \E_{q_\bz} \sbr{\nabla_\bz \sbr{ \log q_\bz\brt}\, h_\bz\brt} \text{.}
\end{align}
Algorithm \ref{alg:BB} sketches the gradient-free optimization approach. Note that at each iteration, the expectation in \eqref{eq:bb_gradient} is approximated with $S$ samples from the posterior $q_{\bz_t}$. Different optimization algorithms differ in how $\bz$ is defined (e.g., it updating a natural parameter), in how natural-gradient computations are performed, and in the adoption of alternatives forms for \eqref{eq:sgd_on_zeta} (e.g., using a retraction in manifold optimization). ML research widely adopts a Gaussian prior of zero mean and covariance matrix $\tau I$, with $\tau>0$.
\vspace{-0.3cm}
\begin{center}
\scalebox{0.9}{
\begin{minipage}{0.9\linewidth}
\begin{algorithm}[H]
\caption{General form of a gradient-free VI optimizer}
\label{alg:BB}
\begin{algorithmic}
\State Set hyperparameters (here $\beta$, $S$, $\tau$), $t = 0$
\State Set initial values $\bz_0$
\Repeat
    \State Simulate $\bth_s \sim q_{\bz_t}$, for $s=1,\dots,S$
    \State $h_{\bz_t}\brt = \log p\brt + \log p\br{y \vert \bth} - \log q_{\bz_t}\brt$
    \State $\hat{\nabla}_{\bz_t} \LB = \frac{1}{S}\sum_{s=1}^{S} \left. \nabla_\bz \log q_\bz\br{\bth_s} \right|_{\bz = \bz_t} \times h_{\bz_t} \br{\bth_s}$
    \State $\bz_{t+1} = \bz_t +\beta \hat{\nabla}_{\bz_t} \LB$
    \State  $t = t+1$
\Until{stopping criterion is met}
\end{algorithmic}
\end{algorithm}
\end{minipage}
}
\end{center}
We briefly introduce the three state-of-the-art optimizers adopted in the empirical analysis.
The gradient in \eqref{eq:bb_gradient} is often called a black-box gradient. 
Despite the terminology, the black-box approach has not to be intended as an opaque mechanism, but as a transparent and accessible solution for computing lower bound's derivatives without explicitly requiring model's derivatives. The expectation in \eqref{eq:bb_gradient}, as of Algorithm \ref{alg:BB}, is computed as an average of products between the easy-to-derive gradients of the variational loglikelihood $\nabla_\bz \log q_{\bz}\br{\bth_s}$ computed in $\bz=\bz_t$ and the $h$-function $h_{\bz_t}\br{\bth_s}$, so that the computation of $\hat{\nabla}_{\bz_t} \LB$ involves only $h$-function's queries, and not its gradients w.r.t. $\bz$.

Black-box VI (BBVI), \citep{ranganath2014black} uses the rule \eqref{eq:sgd_on_zeta} applied to Euclidean black-box gradients, computed as in \eqref{eq:bb_gradient}.
Quasi-Black box VI (QBVI) \citep{magris2022quasi} extends BBVI using natural gradients. QBVI relies on a natural-parameter parametrization of the variational posterior enabling natural gradient updates without requiring the explicit computation and inversion of the Fisher matrix. This is a relevant computational advantage. BBVI and QBVI are broadly applied under a diagonal covariance matrix specification and a log-variance parametrization, as they cannot guarantee the positive definiteness of the variational covariance matrix. Conversely, the two are of low complexity as matrix operations (especially inversion) are straightforward.
Manifold Gaussian Variational Bayes (MGVB) is a black-box approach, boosted by natural gradients, relying on manifold optimization to grant the positive definiteness of the full covariance matrix \cite{tran2021variational}. MGVB solves the positive definiteness issue while allowing for additional modeling flexibility provided by its full covariance specification.
Certain theoretical issues and some approximations that MGVB relies upon are resolved by the Exact Manifold Gaussian Variational Bayes (EMGVB) approach, that further improves the computation of the natural gradients \citep{magris2022exact}.

\begin{figure}[h!]
    \centering
    \includegraphics[width=0.47\textwidth,trim={1cm 0cm 0cm 0.5cm},clip]{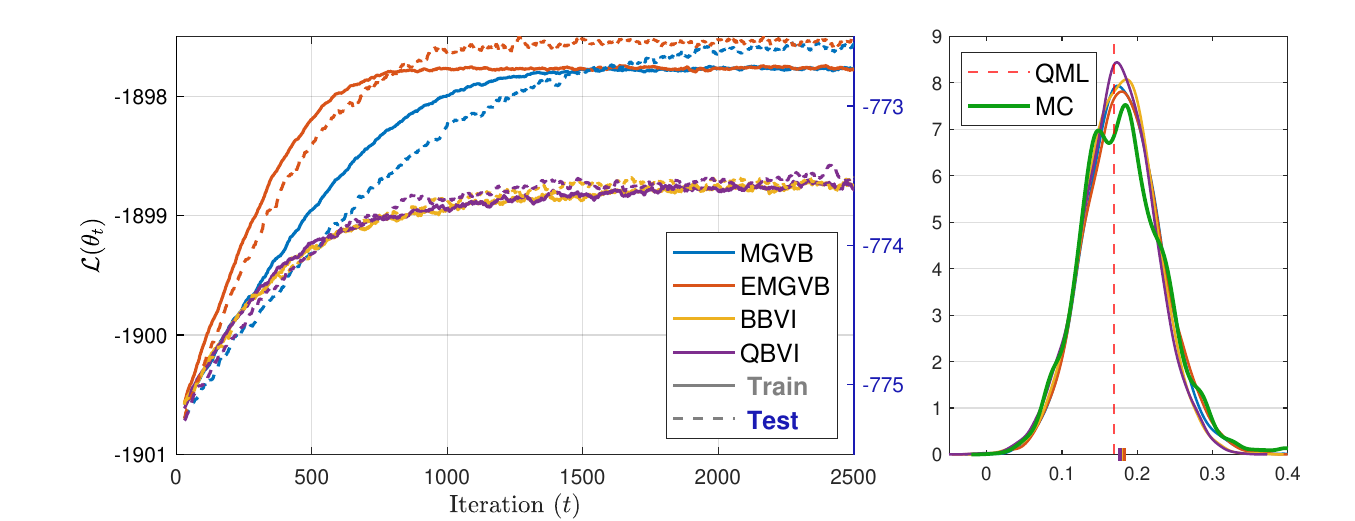}
    \caption{Lower bound optimization (left) and posterior distribution of for the $\gamma$ parameter (right). GJR-GARCH(1,1), Microsoft Inc. data.}
    \label{fig:lb}
\end{figure}

\subsection{Transformations}\label{sec:trans}
It is clear that the application of Gaussian VI is problematic for the heavily constrained volatility models of Sec. \ref{sec:models:garch}. For instance, a Gaussian posterior is incompatible with the $\omega>0$ or the $0<d<1$ requirements and plausibly with a Gaussian covariance structure (Gaussian copula). Moreover, the adoption of Gaussian priors is also inadequate. These issues can be fixed with appropriate parameter transforms.

In Algorithm \ref{alg:BB}, two steps are critical: (i) sampling from the Gaussian variational posterior and (ii) evaluating the model log-likelihood. By adopting the VI Gaussian framework, the unconstrained components of a sample $\bth$ from  $q_{\bz = \{\bmu,\text{vec}(\S)\}}$ need to be appropriately transformed into the valid constrained space for evaluating the log-likelihood. This is done, e.g., in \citep{tran2021variational} for the GARCH(1,1), and aligned with the well-adopted rationale for VI in medium-scale ML models of \citep{kucukelbir2017automatic}.


Let $\bnu$ denote a $d$-dimensional parameter parametrizing a GARCH-family model $m$, and $\mathcal{C}_m$ the constrained parameter space where $\bnu$ lives. Be $\psi_m:\mathcal{C}_m\mapsto \R^d$ a transform that maps $\bnu \in \mathcal{C}_m$ to $\bth \in \R^d$. Let $\psi^{-1}_m$ denote the corresponding inverse transform, i.e., $\bnu = \psi^{-1}_m\br{\bth}$. This is the relevant transform in VI; we will show that such $\psi^{-1}_m$ exists and is of simple form for the models in Sec. \ref{sec:models:garch}. 
Through $\psi^{-1}_m$ we can apply the update \eqref{eq:sgd_on_zeta}, map posterior samples $\bth\sim q_\bz$ into $\mathcal{C}_m$ as $\psi^{-1}_m(\bth)$, evaluate the likelihood, and approximate the expectation in \eqref{eq:bb_gradient}. Similarly, we can, e.g., compare the QML estimates with the mean of transformed posterior samples, interpretable as approximations of the transformed posterior living in $\mathcal{C}_m$:
\begin{equation}\label{eq:post_trans_mean}
\E_{q_{\bz^\star}}\sbr{\psi^{-1}_m (\bth^\star)} \approx \frac{1}{S} \sum_{s=1}^S \psi^{-1}_m(\bth^\star_s),\, \bth^\star_s\sim q_{\bz^\star}\text{.}
\end{equation}
In principle, QML could optimize $\ell(\psi^{-1}_m\brt)$, yet the use of constrained optimization is preponderant (e.g., in Python's \texttt{arch} and R's \texttt{rugarch} packages), so that parameter transforms are not relevant for standard QML estimation. Indeed, we are unaware of any work presenting such transformations. As fundamental for applying VI, and for future reference, we summarize them in the following propositions. For an element $\theta_\lambda$ of the vector $\bth$ representing a certain parameter $\lambda$ of the model $m$, be $\bth_\lambda$ its corresponding element in $\bth$. Let $f$ denote the logistic function.

\begin{prop}[Inverse transforms for the FIGARCH]
 The FIGARCH constraints $\omega>0$, $0\leq d \leq 1$, $0 \leq \phi \leq (1-d)/2$, $0 \leq \beta \leq d+
 \phi$, are satisfied by the following inverse transforms:
\begin{align*}
    \omega = \exp(\theta_\omega), &\qquad
    d = f(\theta_d),\\
    \phi  = f(\theta_\phi)(1-d)/2,&\qquad
    \beta = f(\omega_\beta)\br{\phi +d}.
\end{align*}
\end{prop}
\textit{Proof}. The result follows immediately from \citep[][footnote 19]{baillie1996fractionally}.

\begin{prop}[Inverse transforms for the GJR-GARCH]\label{prop:garch}
The GJR-GARCH(p,o,q) constraints $\omega>0$, $\alpha_i\geq0$, $\sum \gamma_k + \sum\alpha_i\geq0$, $\beta_j\geq 0$, $\sum(\alpha_i + 1/2\gamma_k +\beta_j)<1$, for $j=1,\dots,p$, $k=1,\dots,o$, $i=1,\dots,q$, for the GJR-GARCH(1,1) model are satisfied by the following inverse transforms:
\begin{align*}
\omega = exp\br{\theta_\omega}, &\qquad \alpha = f(\theta_\alpha), \\
\gamma = f(\theta_\gamma)(2(1-\alpha) + \alpha) -\alpha, &\qquad
\beta = f(\theta_\beta)\br{1-\alpha-1/2\gamma}.
\end{align*}
For the general GJR-GARCH$(p,o,q)$ case, inverse transforms can be computed as for Algorithm \ref{alg:GJR}.
\end{prop}
\textit{Proof}. The transform for $\theta_\omega$ is obvious. As $\gamma$ and $\beta$ are still to be determined, the constraints imply that $\alpha$ can lay anywhere in $\sbr{0,1}$, so $\alpha = f(\omega_\alpha)$. It is required that $\gamma+\alpha>0$,
and $\alpha + \beta +1/2\gamma <1$. Yet $\beta$ is to be determined, so $1/2\gamma<1-\alpha$. The two give $-\alpha<\gamma<2(1-\alpha)$: we first map $\gamma$ to $[2(1-\alpha)+\alpha]$ and then shift the interval by $-\alpha$. I.e.,
$\gamma = f(\omega_\gamma)[2(1-\alpha)+\alpha]-\alpha$. Now map $\theta_\beta$ in $1-\alpha-1/2\gamma$, that is $\beta = f(\omega_\beta)(1-\alpha-1/2\gamma).$ With $p\geq0,o\geq0,q\geq0$, the same interval-partitioning reasoning is sequentially repeated.\\

The last proposition applies to the ARCH ($o=p=0$) and GARCH models ($o=0)$ as a special case. Similarly, one can transform the possibly constrained trainable parameters of any postulated distribution for the iid $\{z_i \}$ innovations. For example, for a $GED\br{\lambda}$ distribution $\lambda = 2+\theta_\lambda^2+\epsilon$, where $\epsilon>0$ is a pedestal to grant $\lambda>2$ holds strictly.

\scalebox{0.85}{
\begin{minipage}{1.1\linewidth}
\begin{algorithm}[H]
\caption{Inverse transformation for the $GJR(p,o,q)$}
\label{alg:GJR}
\begin{multicols}{2}
  \begin{algorithmic}[H]
  \State $\omega = \exp\br{\theta_\omega}$
        \State $s = 1-\varepsilon$
      \For{$i=1,\dots,p$}
        \State $\alpha_i = f\br{\theta_{\alpha_i}}s$
        \State $s = 1-s$
      \EndFor
      \For{$i=1,\dots,o$}
        \State $\gamma_i = f\br{\theta_{\gamma_i}}$
        \If{$p\geq i$}
            \State $\gamma_i = \br{2s +\alpha_i}\gamma_i - \alpha_i$
        \Else
            \State $\gamma_i = 2s\gamma_i$
        \EndIf
        \State $s = 1-0.5s$
       \EndFor
       \For{$i=1,\dots,q$}
        \State $\beta_i = f(\theta_{\beta_i})$
        \State $s = 1-s$
      \EndFor
  \end{algorithmic}
  \end{multicols}
    \vspace{-0.3cm}
\end{algorithm}
\end{minipage}
}

\section{Experiments}
\subsection{Data}
For our empirical analyses, we use daily close-to-close log returns for the constituents of the S\&P500 index.
Our data covers 1383 trading days, from 1 January 2018 to 30 June 2023, divided into train and test sets with a 75\%-25\% split following chronological order. We use 488 stocks, since some constituents changed and their time series are incomplete.

\subsection{Models, and optimization}
To assess how satisfactory VI is in volatility modeling, we 
adopt the following volatility models: ARCH(1), GARCH(1,1), GJR-GARCH(1,1), EGARCH(0,1), EGARCH(1,1), GJR-EGARCH(1,1), FIGARCH(1,d,1). The case study on the Microsoft Corp. data additionally includes the GARCH(2,1), EGARCH(2,1), and the FIGARCH(0,d,1). The analyses adopt the BBVI, QBVI, MGVB, and EMGVB optimizers for VI (the first two under a diagonal variational covariance matrix). For the FIGARCH models, we implement \eqref{eq:figarch} by the method of \citep{nielsen2021infinity}.  As a baseline for comparison, we adopt QML estimates and a Monte Carlo Markov Chain Sampler (MC). An MC reference for VI is advisable as it provides a benchmark for highlighting biases and assessing the quality of the Gaussian variational approximation.

For consistency, in VI, we adopt the same set of hyperparameters for all the experiments and optimizers. In particular, we use a learning rate of $0.005$, $50$ MC draws for approximating the expectation \eqref{eq:bb_gradient}, a diagonal normal prior of unit variance and initial values $\bmu_0 = 0$, $\S_0 = 0.1I$. To increase the stability of the learning process, we update the gradients with a momentum factor of $0.4$. 
Both MC and VI algorithms are run for a longer-than-required number of iterations. This avoids tuning the parameter on a case-by-case basis (which is unfeasible with hundreds of stocks) and provides reasonable guarantees that the algorithms converged. For VI, we observe that typically 1500 iterations are sufficient for the LB to reach a plateau (Fig. \ref{fig:lb}), yet we terminate the training after 2500 iterations. Opposed to ML, in statistics overfitting is a major concern in model selection. For example, in maximum likelihood estimation, the dynamic of the likelihood on a (usually nonexistent) test set is ignored. Early-stopping criteria for MC/VI based on the test loss would lead to non-comparability with the fully-in-sample-optimized maximum likelihood estimates. The relevant data and codes for the experiments are available at \url{github.com/mmagris/GARCHVI}.

\subsection{Results}
\subsubsection{General results}
To assess the effectiveness of VI as a Bayesian procedure, we adopt four performance metrics on the training and test sets. For VI, performances are computed as averages of $7,000$ inversely-transformed samples from the estimated variational posterior $q_{\bz^\star}\br{\bth}$. For MC, the last $7,000$ samples of the Markov chain. The metrics are the Negative Log-Likelihood (NLL), the Root Mean Squared Error (RMSE), the Mean Absolute Deviation (MAD), and the Qlik loss \citep{patton2011volatility}. For the last three, as proxies for the observed conditional variances, we adopt squared returns \citep{patton2011volatility}.

\begin{table*}[h!]
  \centering
  \caption{Main estimation results. Averages and respective standard deviations across the S\&P 500 stocks, of the differences between the individual estimators' performance with respect to the QML performance, expressed as a percentage of the QML performance. Performances are evaluated in the transformed posterior mean \eqref{eq:post_trans_mean}.}
\scalebox{0.75}{
    \begin{tabular}{lllllllll}
          & \multicolumn{4}{c}{Train}     & \multicolumn{4}{c}{Test} \\
\cmidrule(lr){2-5} \cmidrule(lr){6-9}            & \multicolumn{1}{c}{RMSE} & \multicolumn{1}{c}{MAD} & \multicolumn{1}{c}{QLIK} & \multicolumn{1}{c}{NLL} & \multicolumn{1}{c}{RMSE} & \multicolumn{1}{c}{MAD} & \multicolumn{1}{c}{QLIK} & \multicolumn{1}{c}{NLL} \\
    \midrule
    \textbf{ARCH(1)} &       &       &       &       &       &       &       &  \\
    MCMC  & +0.257$\pm$35.820 & +0.488$\pm$42.559 & +0.105$\pm$1.306 & +0.045$\pm$0.750 & +0.422$\pm$40.208 & +0.562$\pm$51.238 & +0.128$\pm$18.353 & +0.048$\pm$6.264 \\
    MGVB  & +0.308$\pm$39.794 & +0.555$\pm$41.888 & +0.113$\pm$1.476 & +0.048$\pm$0.745 & +0.525$\pm$41.932 & +0.663$\pm$52.409 & +0.175$\pm$20.910 & +0.066$\pm$7.305 \\
    EMGVB & +0.313$\pm$39.756 & +0.562$\pm$41.596 & +0.113$\pm$1.474 & +0.048$\pm$0.745 & +0.532$\pm$41.673 & +0.670$\pm$52.077 & +0.176$\pm$20.927 & +0.066$\pm$7.304 \\
    BBVI  & +0.313$\pm$39.755 & +0.562$\pm$41.594 & +0.113$\pm$1.474 & +0.048$\pm$0.745 & +0.532$\pm$41.671 & +0.670$\pm$52.075 & +0.176$\pm$20.927 & +0.066$\pm$7.304 \\
    QBVI  & +0.277$\pm$37.496 & +0.532$\pm$40.190 & +0.116$\pm$1.464 & +0.049$\pm$0.744 & +0.502$\pm$37.718 & +0.639$\pm$50.450 & +0.181$\pm$19.112 & +0.067$\pm$7.027 \\
    \midrule
    \textbf{GARCH(1,1)} &       &       &       &       &       &       &       &  \\
    MCMC  & +0.008$\pm$57.547 & +0.033$\pm$126.751 & +0.279$\pm$8.379 & +0.111$\pm$3.371 & +0.378$\pm$45.971 & +0.320$\pm$117.589 & +0.218$\pm$77.296 & +0.080$\pm$30.211 \\
    MGVB  & -0.019$\pm$56.563 & +0.118$\pm$122.241 & +0.265$\pm$7.860 & +0.106$\pm$3.149 & +0.394$\pm$44.033 & +0.431$\pm$112.530 & +0.221$\pm$73.356 & +0.081$\pm$28.555 \\
    EMGVB & -0.016$\pm$56.838 & +0.129$\pm$123.087 & +0.268$\pm$7.888 & +0.107$\pm$3.170 & +0.399$\pm$44.328 & +0.441$\pm$113.424 & +0.223$\pm$73.864 & +0.082$\pm$28.690 \\
    BBVI  & -0.016$\pm$56.843 & +0.129$\pm$123.097 & +0.268$\pm$7.889 & +0.107$\pm$3.171 & +0.399$\pm$44.333 & +0.441$\pm$113.433 & +0.223$\pm$73.869 & +0.082$\pm$28.692 \\
    QBVI  & -0.098$\pm$40.651 & +0.048$\pm$107.369 & +0.225$\pm$5.025 & +0.090$\pm$1.882 & +0.270$\pm$32.590 & +0.355$\pm$97.150 & +0.244$\pm$47.097 & +0.091$\pm$18.423 \\
    \midrule
    \textbf{GJR-GARCH(1,1)} &       &       &       &       &       &       &       &  \\
    MCMC  & +0.216$\pm$136.296 & +1.076$\pm$188.896 & +0.498$\pm$19.800 & +0.198$\pm$8.189 & +1.023$\pm$110.649 & +1.350$\pm$206.975 & +0.207$\pm$155.432 & +0.079$\pm$60.732 \\
    MGVB  & +0.191$\pm$131.465 & +0.889$\pm$186.580 & +0.509$\pm$19.616 & +0.202$\pm$8.006 & +0.959$\pm$105.886 & +1.140$\pm$204.363 & +0.226$\pm$154.253 & +0.084$\pm$59.804 \\
    EMGVB & +0.173$\pm$119.290 & +0.903$\pm$189.253 & +0.513$\pm$20.138 & +0.204$\pm$8.280 & +0.972$\pm$109.063 & +1.151$\pm$207.438 & +0.235$\pm$156.458 & +0.088$\pm$60.738 \\
    BBVI  & +0.173$\pm$119.297 & +0.903$\pm$189.261 & +0.513$\pm$20.140 & +0.204$\pm$8.281 & +0.972$\pm$109.076 & +1.151$\pm$207.446 & +0.235$\pm$156.471 & +0.088$\pm$60.744 \\
    QBVI  & -0.041$\pm$83.022 & +0.727$\pm$154.161 & +0.408$\pm$12.397 & +0.162$\pm$4.942 & +0.580$\pm$66.795 & +0.998$\pm$163.206 & +0.169$\pm$103.345 & +0.066$\pm$40.164 \\
    \midrule
    \textbf{EGARCH(0,1)} &       &       &       &       &       &       &       &  \\
    MCMC  & +0.591$\pm$100.969 & +0.447$\pm$47.094 & +0.099$\pm$0.889 & +0.043$\pm$0.509 & +0.277$\pm$35.295 & +0.278$\pm$26.375 & +0.079$\pm$4.951 & +0.029$\pm$1.776 \\
    MGVB  & +0.704$\pm$119.838 & +0.418$\pm$49.920 & +0.111$\pm$0.935 & +0.048$\pm$0.540 & +0.335$\pm$49.310 & +0.234$\pm$31.721 & +0.111$\pm$9.794 & +0.041$\pm$3.327 \\
    EMGVB & +0.667$\pm$113.577 & +0.400$\pm$50.086 & +0.107$\pm$0.870 & +0.046$\pm$0.524 & +0.320$\pm$47.369 & +0.216$\pm$30.581 & +0.108$\pm$9.811 & +0.040$\pm$3.336 \\
    BBVI  & +0.667$\pm$113.588 & +0.400$\pm$50.089 & +0.107$\pm$0.870 & +0.046$\pm$0.524 & +0.320$\pm$47.373 & +0.216$\pm$30.584 & +0.108$\pm$9.811 & +0.040$\pm$3.336 \\
    QBVI  & +0.591$\pm$99.846 & +0.338$\pm$46.964 & +0.106$\pm$0.864 & +0.046$\pm$0.518 & +0.258$\pm$38.595 & +0.163$\pm$29.894 & +0.104$\pm$10.196 & +0.039$\pm$3.497 \\
    \midrule
    \textbf{EGARCH(1,1)} &       &       &       &       &       &       &       &  \\
    MCMC  & +0.174$\pm$28.495 & +0.450$\pm$33.825 & +0.199$\pm$1.921 & +0.080$\pm$1.076 & +0.244$\pm$28.261 & +0.501$\pm$57.923 & +0.216$\pm$33.674 & +0.079$\pm$12.473 \\
    MGVB  & +0.131$\pm$30.125 & +0.533$\pm$31.338 & +0.198$\pm$3.107 & +0.079$\pm$1.237 & +0.276$\pm$31.356 & +0.609$\pm$54.202 & +0.237$\pm$33.555 & +0.085$\pm$12.823 \\
    EMGVB & +0.455$\pm$68.170 & +0.708$\pm$83.257 & +0.622$\pm$40.371 & +0.240$\pm$13.503 & +0.571$\pm$63.156 & +0.809$\pm$169.018 & +0.625$\pm$97.310 & +0.224$\pm$35.607 \\
    BBVI  & +0.456$\pm$68.614 & +0.724$\pm$83.494 & +0.623$\pm$40.489 & +0.241$\pm$13.556 & +0.577$\pm$63.868 & +0.807$\pm$164.302 & +0.626$\pm$97.604 & +0.225$\pm$35.741 \\
    QBVI  & +0.397$\pm$76.614 & +0.998$\pm$99.027 & +0.748$\pm$49.880 & +0.288$\pm$16.616 & +0.745$\pm$84.246 & +1.375$\pm$178.473 & +0.749$\pm$115.469 & +0.261$\pm$39.564 \\
    \midrule
    \textbf{GJR-EGARCH(1,1)} &       &       &       &       &       &       &       &  \\
    MCMC  & +0.165$\pm$38.646 & +0.785$\pm$44.847 & +0.259$\pm$2.923 & +0.103$\pm$1.644 & +0.406$\pm$47.171 & +0.849$\pm$83.595 & +0.259$\pm$41.236 & +0.094$\pm$14.483 \\
    MGVB  & +0.202$\pm$56.358 & +0.819$\pm$54.540 & +0.310$\pm$9.159 & +0.127$\pm$4.050 & +0.465$\pm$53.650 & +0.929$\pm$112.795 & +0.310$\pm$60.818 & +0.114$\pm$21.561 \\
    EMGVB & +0.846$\pm$107.619 & +1.214$\pm$120.063 & +1.026$\pm$71.257 & +0.394$\pm$24.872 & +1.043$\pm$146.509 & +1.124$\pm$272.713 & +0.781$\pm$161.688 & +0.276$\pm$55.370 \\
    BBVI  & +0.847$\pm$108.432 & +1.235$\pm$120.755 & +1.028$\pm$71.166 & +0.394$\pm$24.853 & +1.047$\pm$146.676 & +1.150$\pm$273.753 & +0.783$\pm$162.340 & +0.277$\pm$55.636 \\
    QBVI  & +0.828$\pm$139.379 & +1.831$\pm$153.957 & +1.170$\pm$79.713 & +0.449$\pm$27.667 & +1.412$\pm$187.324 & +2.177$\pm$324.510 & +0.897$\pm$175.608 & +0.326$\pm$64.718 \\
    \midrule
    \textbf{FIGARCH(1,d,1)} &       &       &       &       &       &       &       &  \\
    MCMC  & +0.090$\pm$54.384 & +0.845$\pm$136.805 & +0.258$\pm$15.436 & +0.103$\pm$6.538 & +0.204$\pm$48.465 & +1.007$\pm$205.486 & -0.028$\pm$64.930 & -0.018$\pm$25.460 \\
    MGVB  & +0.146$\pm$59.358 & +2.381$\pm$217.490 & +0.149$\pm$21.112 & +0.061$\pm$8.566 & +0.475$\pm$67.696 & +2.620$\pm$277.691 & -0.587$\pm$101.008 & -0.213$\pm$36.897 \\
    EMGVB & +0.172$\pm$68.563 & +2.477$\pm$240.598 & +0.153$\pm$21.398 & +0.063$\pm$8.714 & +0.514$\pm$76.643 & +2.762$\pm$307.879 & -0.569$\pm$105.048 & -0.206$\pm$38.216 \\
    BBVI  & +0.172$\pm$68.730 & +2.478$\pm$240.930 & +0.153$\pm$21.404 & +0.063$\pm$8.717 & +0.514$\pm$76.825 & +2.764$\pm$308.238 & -0.569$\pm$105.073 & -0.206$\pm$38.222 \\
    QBVI  & +0.145$\pm$58.516 & +2.427$\pm$217.864 & +0.168$\pm$20.448 & +0.069$\pm$8.299 & +0.493$\pm$67.799 & +2.716$\pm$301.438 & -0.577$\pm$99.190 & -0.209$\pm$36.377 \\
    \bottomrule
    \end{tabular}%
    }
  \label{tab:main}%
\end{table*}%

Table \ref{tab:main} presents the overall estimation results for the 488 stocks. 
For a performance metric $M_E^x$ computed on a data subsample $x$ as a result of an estimation procedure $E$, the entries in the Table correspond to mean performance deviations from the QML benchmark expressed as percentages, i.e., $100( M^x_{E}/M^x_{QML}-1)$, and their standard deviations for the S\&P 500 stocks.
We comment on the predominance of positive signs indicating that QML is, overall, the preferred estimation approach from a merely quantitative perspective. However, VI methods can sometimes outperform QML in certain model/loss combinations.
Clearly, test NLL values are always positive, confirming that QML estimates are optimal in this sense.
At a general level, the typical magnitude of the ratios is in the sub-1\% order, indicating that MC/VI estimation procedures are indeed effective w.r.t. QML and each other.
Among the VI optimizers, we do not observe patterns indicating the dominance of one optimizer over another, implying they all determine a comparable variational approximation and, thus, performance. At the same time, the homogeneity in the results indicates that VI is robust with respect to the choice of the optimizer; all the optimizers appear adequate for the problems analyzed. By applying Chebyshev's inequality with, e.g., a margin of 3 standard deviations, the differences are broadly insignificant, indicating that the Gaussian variational approximation for the unconstrained parameters is plausible, at least for capturing the first two moments of the margins.

Regardless of the VI optimizer and volatility model, these results show that VI firmly stands as a valid estimation alternative to MC sampling. The in-sample and out-of-sample loss in performance w.r.t. QML is negligible, while the Bayesian framework enables several advantages, as for Sec. \ref{sec:introd}.


\begin{figure}[t!]
    \centering
    \includegraphics[width=0.46\textwidth,trim={1.4cm 0.3cm 1.8cm 0.5cm},clip]{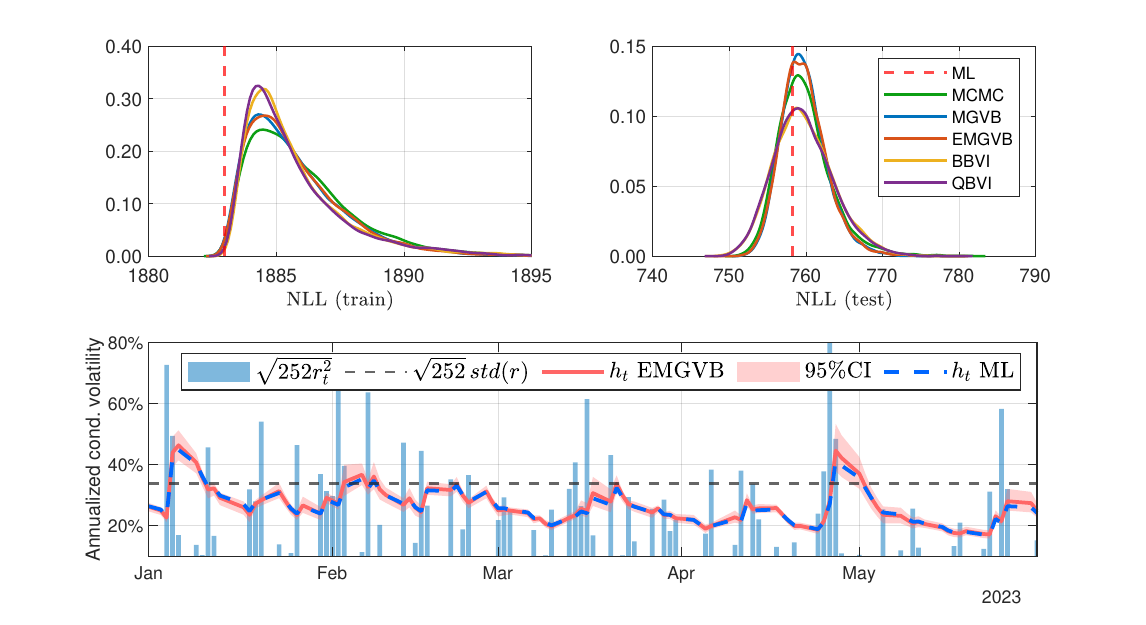}
    \caption{Distribution of the train (top left) and test (top right) NNL, and confidence bounds for the conditional volatility (bottom). GJR-GARCH(1,1), Microsoft Inc. data.}
    \label{fig:NLLcondVar}
\end{figure}

\subsubsection{Case study}
We provide further analyses for Microsoft Corp. data. Table \ref{tab:msft} summarizes the results regarding the training processes and the parameters' estimates.

\begin{table*}
  \centering
  \caption{Estimation results for the Microsoft Inc. data for the GJR-GARCH(1,1). Differences with respect to the QML estimates and standard deviations. Parameter estimates refer to transformed posterior means \eqref{eq:post_trans_mean}, while performances' means and standard deviations are computed from transformed posterior samples.}
  \scalebox{0.67}{
        \begin{tabular}{lcccccccccccccc}
          & \multicolumn{4}{c}{Parameters} & \multicolumn{2}{c}{Train} & \multicolumn{2}{c}{Test} & \multicolumn{3}{c}{Train} & \multicolumn{3}{c}{Test} \\
\cmidrule(lr){2-5}   \cmidrule(lr){6-7}   \cmidrule(lr){8-9}  \cmidrule(lr){10-12} \cmidrule(lr){13-15}        & $\omega$ & $\alpha$ &     &       & $\mathcal{L}(\bf{\theta}^\star)$    & $p(y\vert \bf{\theta}^\star)$   & $\mathcal{L}(\bf{\theta}^\star)$    & $p(y\vert \bf{\theta}^\star)$  & RMSE  & QLIK  & NLL   & RMSE  & QLIK  & NLL \\
    \midrule
    \multicolumn{2}{l}{\textit{\textbf{ARCH(1)}}} &       &       &       &       &       &       &       &       &       &       &       &       &  \\
    ML    & 1.92  & 0.43  &       &       &       & -1966.75 &       & -770.55 & 10.04 & 1.67  & 1966.75 & 8.49  & 1.58  & 770.55 \\
    MCMC  & 1.905$\pm$0.117 & 0.446$\pm$0.066 &       &       &       & -1966.77 &       & -770.64 & 10.040$\pm$0.131 & 1.670$\pm$0.002 & 1967.688$\pm$0.914 & 8.548$\pm$0.193 & 1.579$\pm$0.024 & 770.981$\pm$4.151 \\
    MGVB  & 1.904$\pm$0.121 & 0.446$\pm$0.066 &       &       & -1971.19 & -1966.77 & -774.44 & -770.68 & 10.040$\pm$0.137 & 1.670$\pm$0.002 & 1967.726$\pm$0.930 & 8.549$\pm$0.191 & 1.579$\pm$0.025 & 771.049$\pm$4.281 \\
    EMGVB & 1.904$\pm$0.122 & 0.446$\pm$0.066 &       &       & -1971.19 & -1966.77 & -774.44 & -770.69 & 10.041$\pm$0.138 & 1.670$\pm$0.002 & 1967.731$\pm$0.972 & 8.547$\pm$0.190 & 1.579$\pm$0.025 & 771.043$\pm$4.307 \\
    BBVI  & 1.907$\pm$0.108 & 0.445$\pm$0.057 &       &       & -1971.32 & -1966.77 & -774.53 & -770.60 & 10.036$\pm$0.120 & 1.670$\pm$0.002 & 1967.749$\pm$1.097 & 8.541$\pm$0.161 & 1.579$\pm$0.027 & 771.028$\pm$4.582 \\
    QBVI  & 1.905$\pm$0.108 & 0.446$\pm$0.057 &       &       & -1971.32 & -1966.77 & -774.55 & -770.63 & 10.034$\pm$0.120 & 1.670$\pm$0.002 & 1967.741$\pm$1.100 & 8.545$\pm$0.162 & 1.579$\pm$0.027 & 771.062$\pm$4.557 \\
    \midrule
    \multicolumn{2}{l}{\textit{\textbf{GARCH(1,1)}}} &       & $\beta$ &       &       &       &       &       &       &       &       &       &       &  \\
    ML    & 0.15  & 0.19  & 0.77  &       &       & -1890.44 &       & -761.47 & 10.51 & 1.52  & 1890.44 & 8.09  & 1.52  & 761.42 \\
    MCMC  & 0.200$\pm$0.046 & 0.203$\pm$0.031 & 0.737$\pm$0.036 &       &       & -1890.98 &       & -762.31 & 10.446$\pm$0.072 & 1.523$\pm$0.003 & 1892.321$\pm$1.502 & 8.117$\pm$0.067 & 1.532$\pm$0.015 & 762.932$\pm$2.539 \\
    MGVB  & 0.205$\pm$0.047 & 0.201$\pm$0.030 & 0.736$\pm$0.035 &       & -1901.36 & -1891.06 & -771.91 & -762.49 & 10.444$\pm$0.071 & 1.523$\pm$0.003 & 1892.456$\pm$1.715 & 8.111$\pm$0.061 & 1.532$\pm$0.017 & 762.999$\pm$2.912 \\
    EMGVB & 0.205$\pm$0.047 & 0.201$\pm$0.029 & 0.736$\pm$0.035 &       & -1901.36 & -1891.06 & -771.90 & -762.51 & 10.444$\pm$0.071 & 1.523$\pm$0.003 & 1892.446$\pm$1.698 & 8.111$\pm$0.061 & 1.532$\pm$0.017 & 763.017$\pm$2.910 \\
    BBVI  & 0.196$\pm$0.021 & 0.199$\pm$0.023 & 0.740$\pm$0.023 &       & -1902.27 & -1890.89 & -772.87 & -762.55 & 10.449$\pm$0.048 & 1.523$\pm$0.003 & 1892.334$\pm$1.740 & 8.106$\pm$0.049 & 1.532$\pm$0.022 & 762.995$\pm$3.708 \\
    QBVI  & 0.197$\pm$0.020 & 0.199$\pm$0.023 & 0.740$\pm$0.023 &       & -1902.27 & -1890.92 & -772.88 & -762.60 & 10.449$\pm$0.048 & 1.523$\pm$0.003 & 1892.387$\pm$1.783 & 8.105$\pm$0.049 & 1.532$\pm$0.022 & 762.939$\pm$3.769 \\
    \midrule
    \multicolumn{2}{l}{\textit{\textbf{GARCH(2,1)}}} &       & $\beta_1$ & $\beta_2$ &       &       &       &       &       &       &       &       &       &  \\
    ML    & 0.17  & 0.21  & 0.62  & 0.13  &       & -1890.26 &       & -761.03 & 10.43 & 1.52  & 1890.26 & 8.09  & 1.52  & 760.99 \\
    MCMC  & 0.209$\pm$0.052 & 0.231$\pm$0.038 & 0.545$\pm$0.117 & 0.160$\pm$0.092 &       & -1890.65 &       & -762.61 & 10.372$\pm$0.086 & 1.522$\pm$0.003 & 1892.089$\pm$1.381 & 8.140$\pm$0.077 & 1.527$\pm$0.015 & 762.110$\pm$2.589 \\
    MGVB  & 0.212$\pm$0.051 & 0.229$\pm$0.035 & 0.543$\pm$0.100 & 0.164$\pm$0.080 & -1898.50 & -1890.59 & -768.27 & -762.10 & 10.372$\pm$0.084 & 1.522$\pm$0.003 & 1892.104$\pm$1.566 & 8.134$\pm$0.071 & 1.526$\pm$0.018 & 761.916$\pm$3.031 \\
    EMGVB & 0.211$\pm$0.053 & 0.230$\pm$0.035 & 0.543$\pm$0.103 & 0.164$\pm$0.083 & -1898.49 & -1890.59 & -768.27 & -762.12 & 10.371$\pm$0.085 & 1.522$\pm$0.003 & 1892.127$\pm$1.581 & 8.135$\pm$0.071 & 1.526$\pm$0.018 & 761.918$\pm$3.010 \\
    BBVI  & 0.204$\pm$0.022 & 0.229$\pm$0.025 & 0.555$\pm$0.047 & 0.159$\pm$0.034 & -1900.28 & -1890.50 & -769.82 & -761.34 & 10.372$\pm$0.056 & 1.523$\pm$0.005 & 1892.379$\pm$2.315 & 8.134$\pm$0.055 & 1.526$\pm$0.026 & 761.997$\pm$4.391 \\
    QBVI  & 0.205$\pm$0.022 & 0.230$\pm$0.024 & 0.552$\pm$0.047 & 0.161$\pm$0.034 & -1900.27 & -1890.50 & -769.85 & -761.33 & 10.370$\pm$0.056 & 1.523$\pm$0.005 & 1892.384$\pm$2.329 & 8.134$\pm$0.053 & 1.526$\pm$0.026 & 762.071$\pm$4.394 \\
    \midrule
    \multicolumn{2}{l}{\textit{\textbf{GJR-GARCH(1,1)}}} &       & $\gamma$ & $\beta$ &       &       &       &       &       &       &       &       &       &  \\
    ML    & 0.17  & 0.09  & 0.17  & 0.77  &       & -1882.96 &       & -758.24 & 10.59 & 1.50  & 1882.96 & 8.02  & 1.50  & 758.18 \\
    MCMC  & 0.218$\pm$0.046 & 0.111$\pm$0.031 & 0.182$\pm$0.056 & 0.731$\pm$0.034 &       & -1883.76 &       & -759.25 & 10.560$\pm$0.071 & 1.510$\pm$0.004 & 1885.662$\pm$1.819 & 8.089$\pm$0.075 & 1.517$\pm$0.017 & 760.414$\pm$2.902 \\
    MGVB  & 0.218$\pm$0.045 & 0.111$\pm$0.031 & 0.182$\pm$0.050 & 0.731$\pm$0.035 & -1897.75 & -1883.77 & -772.49 & -759.43 & 10.556$\pm$0.060 & 1.509$\pm$0.004 & 1885.575$\pm$1.906 & 8.088$\pm$0.070 & 1.516$\pm$0.018 & 760.344$\pm$3.151 \\
    EMGVB & 0.218$\pm$0.047 & 0.111$\pm$0.030 & 0.182$\pm$0.051 & 0.730$\pm$0.035 & -1897.74 & -1883.77 & -772.48 & -759.42 & 10.557$\pm$0.061 & 1.510$\pm$0.004 & 1885.590$\pm$1.889 & 8.088$\pm$0.070 & 1.516$\pm$0.018 & 760.350$\pm$3.081 \\
    BBVI  & 0.212$\pm$0.021 & 0.112$\pm$0.028 & 0.178$\pm$0.048 & 0.733$\pm$0.022 & -1898.68 & -1883.65 & -773.50 & -759.57 & 10.550$\pm$0.051 & 1.509$\pm$0.004 & 1885.548$\pm$1.997 & 8.080$\pm$0.052 & 1.516$\pm$0.024 & 760.368$\pm$4.023 \\
    QBVI  & 0.209$\pm$0.020 & 0.111$\pm$0.029 & 0.177$\pm$0.049 & 0.735$\pm$0.022 & -1898.69 & -1883.60 & -773.42 & -759.44 & 10.551$\pm$0.052 & 1.509$\pm$0.004 & 1885.490$\pm$1.985 & 8.078$\pm$0.051 & 1.516$\pm$0.024 & 760.263$\pm$4.038 \\
    \midrule
    \multicolumn{2}{l}{\textit{\textbf{EGARCH(0,1)}}} &       &       &       &       &       &       &       &       &       &       &       &       &  \\
    ML    & 1.12  & 0.50  &       &       &       & -1999.77 &       & -758.82 & 11.68 & 1.73  & 1999.77 & 8.07  & 1.51  & 758.82 \\
    MCMC  & 1.121$\pm$0.052 & 0.502$\pm$0.050 &       &       &       & -1999.77 &       & -758.86 & 11.694$\pm$0.025 & 1.735$\pm$0.002 & 2000.717$\pm$0.916 & 8.091$\pm$0.081 & 1.510$\pm$0.020 & 759.203$\pm$3.378 \\
    MGVB  & 1.121$\pm$0.053 & 0.501$\pm$0.050 &       &       & -2006.49 & -1999.77 & -764.15 & -758.85 & 11.695$\pm$0.027 & 1.735$\pm$0.002 & 2000.757$\pm$0.972 & 8.091$\pm$0.081 & 1.510$\pm$0.021 & 759.222$\pm$3.511 \\
    EMGVB & 1.121$\pm$0.053 & 0.501$\pm$0.050 &       &       & -2006.49 & -1999.77 & -764.23 & -758.85 & 11.694$\pm$0.027 & 1.735$\pm$0.002 & 2000.745$\pm$0.996 & 8.090$\pm$0.080 & 1.510$\pm$0.020 & 759.194$\pm$3.493 \\
    BBVI  & 1.120$\pm$0.051 & 0.500$\pm$0.048 &       &       & -2006.52 & -1999.77 & -764.31 & -758.87 & 11.694$\pm$0.024 & 1.735$\pm$0.002 & 2000.761$\pm$1.039 & 8.088$\pm$0.080 & 1.510$\pm$0.022 & 759.202$\pm$3.780 \\
    QBVI  & 1.120$\pm$0.051 & 0.500$\pm$0.048 &       &       & -2006.52 & -1999.77 & -764.51 & -758.87 & 11.694$\pm$0.024 & 1.735$\pm$0.002 & 2000.765$\pm$1.050 & 8.087$\pm$0.080 & 1.510$\pm$0.023 & 759.209$\pm$3.837 \\
    \midrule
    \multicolumn{2}{l}{\textit{\textbf{EGARCH(1,1)}}} &       & $\beta$ &       &       &       &       &       &       &       &       &       &       &  \\
    ML    & 0.07  & 0.35  & 0.94  &       &       & -1893.86 &       & -754.74 & 10.49 & 1.53  & 1893.86 & 7.97  & 1.48  & 754.74 \\
    MCMC  & 0.076$\pm$0.018 & 0.371$\pm$0.044 & 0.933$\pm$0.016 &       &       & -1893.95 &       & -755.12 & 10.480$\pm$0.105 & 1.529$\pm$0.002 & 1895.328$\pm$1.243 & 8.001$\pm$0.060 & 1.487$\pm$0.012 & 755.318$\pm$2.055 \\
    MGVB  & 0.076$\pm$0.017 & 0.368$\pm$0.044 & 0.933$\pm$0.016 &       & -1906.51 & -1893.94 & -766.16 & -755.04 & 10.488$\pm$0.110 & 1.529$\pm$0.003 & 1895.409$\pm$1.287 & 7.997$\pm$0.057 & 1.487$\pm$0.012 & 755.263$\pm$2.128 \\
    EMGVB & 0.082$\pm$0.021 & 0.382$\pm$0.047 & 0.927$\pm$0.019 &       & -1906.63 & -1894.18 & -766.23 & -755.34 & 10.478$\pm$0.122 & 1.530$\pm$0.003 & 1895.999$\pm$1.689 & 8.012$\pm$0.065 & 1.489$\pm$0.014 & 755.647$\pm$2.423 \\
    BBVI  & 0.093$\pm$0.011 & 0.399$\pm$0.039 & 0.917$\pm$0.010 &       & -1907.75 & -1894.75 & -767.06 & -755.52 & 10.461$\pm$0.115 & 1.531$\pm$0.003 & 1896.368$\pm$1.762 & 8.024$\pm$0.061 & 1.490$\pm$0.018 & 755.929$\pm$3.039 \\
    QBVI  & 0.096$\pm$0.011 & 0.403$\pm$0.039 & 0.915$\pm$0.011 &       & -1907.92 & -1894.94 & -767.06 & -755.55 & 10.457$\pm$0.117 & 1.531$\pm$0.004 & 1896.681$\pm$1.945 & 8.029$\pm$0.063 & 1.490$\pm$0.019 & 755.898$\pm$3.170 \\
    \midrule
    \multicolumn{2}{l}{\textit{\textbf{EGARCH(2,1)}}} &       & $\beta_1$ & $\beta_2$ &       &       &       &       &       &       &       &       &       &  \\
    ML    & 0.07  & 0.38  & 0.85  & 0.09  &       & -1893.73 &       & -754.23 & 10.47 & 1.53  & 1893.73 & 7.97  & 1.48  & 754.23 \\
    MCMC  & 0.079$\pm$0.020 & 0.395$\pm$0.058 & 0.813$\pm$0.152 & 0.117$\pm$0.147 &       & -1893.81 &       & -754.54 & 10.483$\pm$0.115 & 1.529$\pm$0.003 & 1895.766$\pm$1.553 & 8.008$\pm$0.077 & 1.486$\pm$0.014 & 755.107$\pm$2.330 \\
    MGVB  & 0.079$\pm$0.018 & 0.392$\pm$0.053 & 0.813$\pm$0.141 & 0.118$\pm$0.138 & -1908.04 & -1893.80 & -767.03 & -754.29 & 10.485$\pm$0.115 & 1.529$\pm$0.003 & 1895.665$\pm$1.459 & 8.002$\pm$0.063 & 1.484$\pm$0.013 & 754.826$\pm$2.281 \\
    EMGVB & 0.093$\pm$0.025 & 0.419$\pm$0.055 & 0.799$\pm$0.103 & 0.118$\pm$0.099 & -1908.55 & -1894.32 & -767.41 & -755.00 & 10.466$\pm$0.136 & 1.531$\pm$0.004 & 1896.708$\pm$2.066 & 8.029$\pm$0.077 & 1.488$\pm$0.016 & 755.544$\pm$2.774 \\
    BBVI  & 0.111$\pm$0.013 & 0.443$\pm$0.043 & 0.829$\pm$0.013 & 0.074$\pm$0.012 & -1912.81 & -1895.66 & -770.53 & -755.37 & 10.431$\pm$0.138 & 1.534$\pm$0.006 & 1898.215$\pm$3.089 & 8.063$\pm$0.083 & 1.491$\pm$0.023 & 756.064$\pm$3.852 \\
    QBVI  & 0.111$\pm$0.013 & 0.447$\pm$0.043 & 0.817$\pm$0.014 & 0.085$\pm$0.013 & -1912.87 & -1895.68 & -770.48 & -755.45 & 10.433$\pm$0.140 & 1.535$\pm$0.007 & 1898.410$\pm$3.372 & 8.065$\pm$0.081 & 1.493$\pm$0.024 & 756.304$\pm$4.079 \\
    \midrule
    \multicolumn{2}{l}{\textit{\textbf{GJR-EGARCH(1,1)}}} &       & $\gamma$ & $\beta$ &       &       &       &       &       &       &       &       &       &  \\
    ML    & 0.07  & 0.30  & -0.10 & 0.94  &       & -1885.35 &       & -749.47 & 10.52 & 1.51  & 1885.35 & 7.91  & 1.45  & 749.47 \\
    MCMC  & 0.073$\pm$0.016 & 0.319$\pm$0.047 & -0.105$\pm$0.025 & 0.931$\pm$0.015 &       & -1885.47 &       & -750.24 & 10.526$\pm$0.105 & 1.513$\pm$0.003 & 1887.398$\pm$1.465 & 7.945$\pm$0.072 & 1.460$\pm$0.014 & 750.668$\pm$2.437 \\
    MGVB  & 0.074$\pm$0.016 & 0.320$\pm$0.045 & -0.106$\pm$0.026 & 0.930$\pm$0.015 & -1901.83 & -1885.47 & -764.69 & -750.01 & 10.521$\pm$0.102 & 1.513$\pm$0.003 & 1887.407$\pm$1.454 & 7.944$\pm$0.061 & 1.458$\pm$0.014 & 750.435$\pm$2.368 \\
    EMGVB & 0.084$\pm$0.021 & 0.342$\pm$0.049 & -0.108$\pm$0.031 & 0.921$\pm$0.019 & -1902.11 & -1885.95 & -764.90 & -750.65 & 10.514$\pm$0.112 & 1.515$\pm$0.004 & 1888.489$\pm$1.990 & 7.977$\pm$0.079 & 1.463$\pm$0.016 & 751.266$\pm$2.764 \\
    BBVI  & 0.095$\pm$0.010 & 0.357$\pm$0.039 & -0.113$\pm$0.028 & 0.911$\pm$0.010 & -1903.33 & -1886.69 & -766.01 & -750.98 & 10.511$\pm$0.109 & 1.516$\pm$0.004 & 1888.890$\pm$2.033 & 7.998$\pm$0.081 & 1.465$\pm$0.019 & 751.582$\pm$3.324 \\
    QBVI  & 0.098$\pm$0.011 & 0.363$\pm$0.039 & -0.114$\pm$0.028 & 0.908$\pm$0.011 & -1903.58 & -1886.97 & -765.99 & -751.10 & 10.511$\pm$0.111 & 1.517$\pm$0.004 & 1889.281$\pm$2.176 & 8.009$\pm$0.087 & 1.466$\pm$0.021 & 751.790$\pm$3.531 \\
    \midrule
    \multicolumn{2}{l}{\boldmath{}\textit{\textbf{FIGARCH(0,$d$,1)}}\unboldmath{}} & $d$   & $\beta$ &       &       &       &       &       &       &       &       &       &       &  \\
    ML    & 0.28  & 0.47  & 0.25  &       &       & -1891.66 &       & -759.11 & 10.40 & 1.52  & 1891.66 & 8.15  & 1.51  & 759.11 \\
    MCMC  & 0.303$\pm$0.086 & 0.511$\pm$0.115 & 0.275$\pm$0.133 &       &       & -1891.64 &       & -759.08 & 10.447$\pm$0.190 & 1.524$\pm$0.002 & 1892.831$\pm$1.070 & 8.180$\pm$0.080 & 1.511$\pm$0.010 & 759.519$\pm$1.718 \\
    MGVB  & 0.340$\pm$0.083 & 0.523$\pm$0.094 & 0.290$\pm$0.109 &       & -1895.55 & -1891.54 & -761.56 & -758.34 & 10.454$\pm$0.174 & 1.524$\pm$0.002 & 1892.737$\pm$1.164 & 8.193$\pm$0.081 & 1.507$\pm$0.010 & 758.735$\pm$1.628 \\
    EMGVB & 0.340$\pm$0.081 & 0.523$\pm$0.094 & 0.291$\pm$0.110 &       & -1895.55 & -1891.54 & -761.57 & -758.34 & 10.454$\pm$0.176 & 1.523$\pm$0.002 & 1892.735$\pm$1.149 & 8.193$\pm$0.081 & 1.507$\pm$0.010 & 758.727$\pm$1.622 \\
    BBVI  & 0.333$\pm$0.072 & 0.493$\pm$0.058 & 0.266$\pm$0.054 &       & -1896.15 & -1891.54 & -761.65 & -758.10 & 10.421$\pm$0.136 & 1.524$\pm$0.003 & 1892.916$\pm$1.548 & 8.182$\pm$0.093 & 1.505$\pm$0.011 & 758.442$\pm$1.921 \\
    QBVI  & 0.330$\pm$0.072 & 0.490$\pm$0.057 & 0.262$\pm$0.053 &       & -1896.15 & -1891.54 & -761.66 & -758.15 & 10.419$\pm$0.135 & 1.524$\pm$0.003 & 1892.944$\pm$1.550 & 8.179$\pm$0.093 & 1.505$\pm$0.011 & 758.444$\pm$1.952 \\
    \midrule
    \multicolumn{2}{l}{\boldmath{}\textit{\textbf{FIGARCH(1,$d$,1)}}\unboldmath{}} & $\phi$ & $d$   & $\beta$ &       &       &       &       &       &       &       &       &       &  \\
    ML    & 0.30  & 0.15  & 0.58  & 0.50  &       & -1890.83 &       & -758.83 & 10.41 & 1.52  & 1890.83 & 8.17  & 1.51  & 758.83 \\
    MCMC  & 0.307$\pm$0.091 & 0.112$\pm$0.055 & 0.549$\pm$0.124 & 0.412$\pm$0.122 &       & -1891.06 &       & -758.97 & 10.395$\pm$0.173 & 1.523$\pm$0.002 & 1892.431$\pm$1.245 & 8.191$\pm$0.083 & 1.511$\pm$0.010 & 759.438$\pm$1.723 \\
    MGVB  & 0.343$\pm$0.082 & 0.108$\pm$0.048 & 0.573$\pm$0.106 & 0.438$\pm$0.104 & -1895.36 & -1890.93 & -761.87 & -758.37 & 10.427$\pm$0.158 & 1.522$\pm$0.002 & 1892.201$\pm$1.209 & 8.197$\pm$0.083 & 1.507$\pm$0.010 & 758.759$\pm$1.657 \\
    EMGVB & 0.343$\pm$0.083 & 0.107$\pm$0.049 & 0.573$\pm$0.108 & 0.437$\pm$0.105 & -1895.35 & -1890.93 & -761.85 & -758.37 & 10.426$\pm$0.159 & 1.523$\pm$0.002 & 1892.232$\pm$1.211 & 8.198$\pm$0.082 & 1.507$\pm$0.010 & 758.763$\pm$1.645 \\
    BBVI  & 0.341$\pm$0.070 & 0.112$\pm$0.047 & 0.567$\pm$0.075 & 0.444$\pm$0.061 & -1895.88 & -1890.89 & -761.95 & -758.14 & 10.435$\pm$0.122 & 1.523$\pm$0.003 & 1892.388$\pm$1.651 & 8.190$\pm$0.092 & 1.505$\pm$0.012 & 758.449$\pm$2.056 \\
    QBVI  & 0.341$\pm$0.071 & 0.113$\pm$0.048 & 0.560$\pm$0.073 & 0.437$\pm$0.060 & -1895.89 & -1890.91 & -761.92 & -758.09 & 10.426$\pm$0.123 & 1.523$\pm$0.003 & 1892.368$\pm$1.561 & 8.188$\pm$0.090 & 1.505$\pm$0.012 & 758.401$\pm$2.016 \\
    \bottomrule
    \end{tabular}%
    }
  \label{tab:msft}%
\end{table*}%

The takeaway, justified blow, is that
differences in performance metrics are broadly negligible, and the differences in the estimated parameters are minor. The impact of the chosen estimation method is secondary, promoting VI as a solid alternative to MC and QML. Performance metrics are broadly overlapping, and differences are not statistically different except for the train NLL, consistently minimized by QML.

It is instructive to look at the performance metrics achieved by the different optimizers.
For all the models, the MSFT findings align with the percentage reported in Table \ref{tab:main}. On both the training and test sets, the values of the performance metrics are remarkably aligned between QML and the Bayesian estimators, also for the additional models. 
Switching to a Bayesian framework does not harm with respect to QML performance on both sets. 
The estimated variance indicates broad non-significance in the difference across the Bayesian estimates, if not for $NLL^\text{train}$. We do not include the additional table with all the cross-testing results but rather visually discuss this case in Figure \ref{fig:NLLcondVar}. The top-right panel of Figure \ref{fig:NLLcondVar} shows that the hypothesis $NLL_{QML}^\text{train} = NLL_{VI}^\text{train}$ cannot be rejected. Conversely, its rejection on the training set validates the above.

Extending the analysis to the value of the optimized LB, we observe that all the VI optimizers are rather equivalent for Bayesian inference, targeting a similar optimum. Clearly, the diagonal BBVI and QBVI ones do not reach the same LB optimum that MGVB and EMGVB do (see. e.g., the GJR-GARCH(1,1) case in Table \ref{tab:msft} and Figure \ref{fig:lb}), yet the differences are well within $1\%$, both on the training and test sets. The differences in the LB correspond to differences in the posterior estimates, explaining differences in the estimates posterior means of Table \ref{tab:msft}. Yet, performance metrics are practically analogous; all the reported estimates can be considered equally effective, especially for out-of-sample forecasting.

In this regard, we observe a remarkable alignment between the MC and MGVB/EMGVB estimates and the standard deviations, suggesting that the full-covariance Gaussian specification appears feasible, at least for capturing the first two moments of the marginal distribution of the true posterior (approximated by the MC sampler).
Figure \ref{fig:lb} includes the posterior means for the GJR-GARCH(1,1) model in the constrained parameter space. The plot highlights the importance of allowing for a full-covariance specification, and the MGVB/EMGVB overlap to the MC density supports the Gaussian variational framework. Observing the QML estimates within the region of high density further validates the overall VI calibration with respect to QML.

As an example of how VI could provide additional insights with respect to the standard frequentist approach, the bottom panel of Figure \ref{fig:NLLcondVar} shows confidence bounds for the predicted conditional variance for 2023.
Enabling a probabilistic dimension for the conditional variance can certainly benefit financial practitioners \citep[][Ch. 6]{ardia2008financial}). For instance, by enabling statistical testing (e.g., for tomorrow's variance being significantly higher than today's), or improving value-at-risk density evaluations, as for Sec. \ref{sec:introd}. Our results indicate that VI can serve this scope well.

\subsection{Robustness checks}
Tables \ref{tab:dist} and \ref{tab:prior} further validate our study by addressing the effect different distributions for the iid innovations $\{z_t\}$ and prior variances have on the estimates.
The reported results involve the MGVB optimizer, the GJR-GARCH(1,1) model, and at the estimated parameter only. In Table \ref{tab:dist}, we consider four distributions and include the corresponding parameters as trainable parameters. The results across the sections of Table \ref{tab:dist} are consistent and robust. The remarkably small ratios in Table \ref{tab:main} are perhaps not related to the choice of the innovations' distribution but rather to the robustness of the VI setup in this setting.

Small prior variance does affect the posterior estimation by keeping the posterior mean rather away from the QML estimates (see Table \ref{tab:prior}). This is certainly positive if one has motivated prior beliefs on the parameter in the unconstrained space (though unlikely). On the other hand, prior variances greater than one already deliver similar estimates (further aligned with QML). Encoding prior lack of knowledge appears to be relatively smooth; a prior variance $\tau I$ with $\tau>1$ is effective in this regard.

\begin{table}[h!]
  \centering
  \caption{Estimation results under different parametric forms of the iid innovations. Microsoft Inc. data, GJR-GARCH(1,1).}
  \scalebox{0.73}{
    \begin{tabular}{clcccccc}
    Error & \multicolumn{1}{c}{Optimizer} & $\omega$ & $\alpha$ & $\gamma$ & $\beta$ & $\nu$ & $\lambda$ \\
    \midrule
    \multirow{3}[2]{*}{Normal} & ML    & 0.169 & 0.089 & 0.169 & 0.772 &       &  \\
          & MCMC  & 0.218 & 0.111 & 0.182 & 0.731 &       &  \\
          & EMGVB & 0.218 & 0.111 & 0.182 & 0.730 &       &  \\
    \midrule
    \multirow{3}[2]{*}{Student-t} & ML    & 0.148 & 0.067 & 0.203 & 0.788 & 6.991 &  \\
          & MCMC  & 0.236 & 0.106 & 0.226 & 0.720 & 6.417 &  \\
          & EMGVB & 0.233 & 0.106 & 0.225 & 0.721 & 6.394 &  \\
    \midrule
    \multirow{3}[2]{*}{GED} & ML    & 0.155 & 0.076 & 0.185 & 0.783 & 1.408 &  \\
          & MCMC  & 0.221 & 0.109 & 0.198 & 0.724 & 1.462 &  \\
          & EMGVB & 0.223 & 0.109 & 0.199 & 0.723 & 1.458 &  \\
    \midrule
    \multirow{3}[2]{*}{Skew-t} & ML    & 0.151 & 0.061 & 0.205 & 0.790 & 7.402 & -0.130 \\
          & MCMC  & 0.230 & 0.102 & 0.227 & 0.725 & 6.521 & -0.120 \\
          & EMGVB & 0.229 & 0.099 & 0.225 & 0.727 & 6.632 & -0.124 \\
    \bottomrule
    \end{tabular}%
    }
  \label{tab:dist}%
\end{table}%

\begin{table}[h!]
  \centering
  \caption{Effect of different prior variances. Microsoft Inc. data,  GJR-GARCH(1,1).}
  \scalebox{0.73}{
    \begin{tabular}{clcccc}
    $\tau$ & \multicolumn{1}{c}{Optimizer} & $\omega$ & $\alpha$ & $\gamma$ & $\beta$ \\
    \midrule
    \multirow{2}[2]{*}{0.01} & MCMC  & 1.033 & 0.477 & 0.242 & 0.210 \\
          & EMGVB & 1.034 & 0.475 & 0.244 & 0.210 \\
    \midrule
    \multirow{2}[2]{*}{0.1} & MCMC  & 0.646 & 0.286 & 0.207 & 0.440 \\
          & EMGVB & 0.650 & 0.285 & 0.212 & 0.439 \\
    \midrule
    \multirow{2}[2]{*}{1} & MCMC  & 0.218 & 0.111 & 0.182 & 0.731 \\
          & EMGVB & 0.218 & 0.111 & 0.182 & 0.730 \\
    \midrule
    \multirow{2}[2]{*}{5} & MCMC  & 0.168 & 0.092 & 0.176 & 0.773 \\
          & EMGVB & 0.173 & 0.092 & 0.176 & 0.769 \\
    \midrule
    \multirow{2}[2]{*}{10} & MCMC  & 0.164 & 0.091 & 0.176 & 0.776 \\
          & EMGVB & 0.166 & 0.089 & 0.175 & 0.775 \\
    \midrule
    \multirow{2}[2]{*}{20} & MCMC  & 0.160 & 0.092 & 0.174 & 0.780 \\
          & EMGVB & 0.163 & 0.088 & 0.175 & 0.778 \\
    \midrule
          & ML    & 0.169 & 0.089 & 0.169 & 0.772 \\
    \bottomrule
    \end{tabular}%
    }
  \label{tab:prior}%
\end{table}%

\section{Conclusion}
This paper documents the validity of Variational Inference (VI) as a tool for the Bayesian estimation for common volatility models of the GARCH family.
We show that within a Gaussian variational framework, VI gradient-free black-box methods are robust and aligned with both the estimates obtained via Monte Carlo sampling and traditional Quasi Maximum Likelihood (QML). In this setting, we show how to adopt parameter transforms to enable VI principles and provide valuable insights on VI by the use of extensive performance statistics calculated from the individual time series of the S\&P500 constituents. Along with a case study and different robustness analyses, we conclude that VI stands as a reliable, adequate, and suitable alternative to MC sampling and QML. The differences in training and test performance metrics with respect to QML and MCMC are typically within the order of $1\%$.
Despite our evidence on the validity of Gaussian variational margins, future research may investigate the appropriateness of the Gaussian copula and the use of alternative dependence structures. More in general, the VI framework could be applied to other domains, such as, e.g., stochastic volatility models or derivative pricing.
We hope that our results will promote the deployment of VI in econometric and financial applications, encouraging the use of further toolsets and results from the Machine Learning research.

\bibliographystyle{plainnat}
\bibliography{Main.bib}

\end{document}